\documentclass[jifs]{iosart2c}

\usepackage{epsfig}
\usepackage{comment}
\usepackage{booktabs}
\usepackage{amsmath,amssymb,amsthm,amsxtra}
\usepackage{amsfonts}
\usepackage{graphicx}
\usepackage{fancyhdr}   
\usepackage{algorithm}	
\usepackage{algorithmic}
\usepackage{subfigure}
\usepackage[T1]{fontenc}
\usepackage{times}%
\usepackage{natbib}
\usepackage[T1]{fontenc}
\usepackage{times}%
\usepackage{url}

\usepackage{natbib}
\usepackage{amsmath}
\usepackage{dcolumn}
\usepackage{graphics}

\newcolumntype{d}[1]{D{.}{.}{#1}}

\firstpage{0} \lastpage{0} \volume{0} \pubyear{2016}

\begin{document}
\begin{frontmatter}                           

%
\title{Drift Robust Non-rigid Optical Flow Enhancement for Long Sequences}

\runningtitle{Drift Robust Non-rigid Optical Flow Enhancement for Long Sequences}

\author[A]{\fnms{Wenbin} \snm{Li}},
\author[B]{\fnms{Darren} \snm{Cosker}}
and
\author[C]{\fnms{Matthew} \snm{Brown}}
\runningauthor{Li et al.}
\address[A]{Department of Computer Science, University College London, WC1E 6BT, UK\\E-mail: w.li@cs.ucl.ac.uk}
\address[B]{Centre for the Analysis of Motion, Entertainment Research and Applications (CAMERA),\\ University of Bath, BA2 7AZ, UK. E-mail: d.p.cosker@bath.ac.uk}
\address[C]{Google, Mountain View, CA 94043, US\\E-mail: m.brown@bath.ac.uk}

\vspace{-10mm}

\begin{abstract}
It is hard to densely track a nonrigid object in long term, which is a fundamental research issue in the computer vision community. This task often relies on estimating pairwise correspondences between images over time where the error is accumulated and leads to a \emph{drift}. In this paper, we introduce a novel optimisation framework with an \emph{Anchor Patch} constraint. It is supposed to significantly reduce overall errors given long sequences containing nonrigidly deformable objects. Our framework can be applied to any dense tracking algorithm, e.g. optical flow. We demonstrate the success of our approach by showing significant error reduction on 6 popular optical flow algorithms applied to a range of realworld nonrigid benchmarks. We also provide quantitative analysis of our approach given synthetic occlusions and image noise.

\end{abstract}


\end{frontmatter}

\section{Introduction}
\label{APO:sec:intro}

Tracking a set of landmark points through multiple images is a fundamental research issue in computer vision. We define tracking in this work as the estimation of corresponding sets of vertices, pixels or landmark points between a reference frame and any other frame in the same image sequence. In the last two decades, optical flow has become a popular approach for tracking through image sequences~\cite{DeCarlo,Borshukov,LME,moBlur,tang} in fields~\cite{lv2014multimodal,lv2013game,reflection,yang2015quality}. Compared with feature matching methods e.g.~\cite{SIFT}, optical flow provides subpixel accuracy and dense correspondence between a pair of images. In this work, we focus in particular on improving tracking in image sequences using optical flow, and our contribution applies to this class of algorithm.

\begin{figure}[t!]
\begin{center}
\includegraphics[width=1\linewidth]{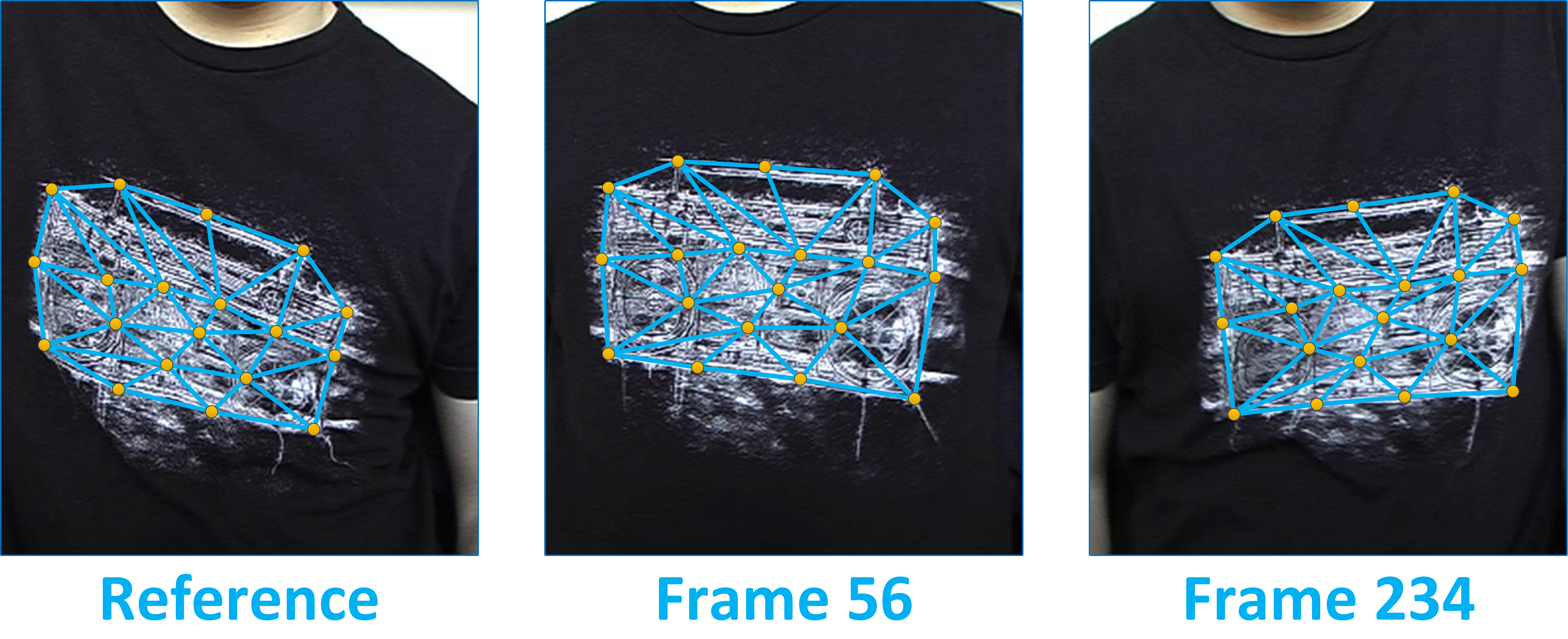}
\end{center}
\vspace{-0.5cm}
\caption{A mesh is tracked through a long nonrigid sequence by using our method.}
\label{APO:fig:firstFig}
\end{figure}

One of the main drawbacks of optical flow is \emph{drift}~\cite{Brox_Large,APO}. Errors accumulated between frames over time result in movement away from the correct tracking trajectory. Between single image pairs, this problem may not be noticeable. However, accumulation when tracking across long sequences can be particularly problematic. Several authors have previously attempted to reduce optical flow \emph{drift} in tracking. DeCarlo \emph{et al.}~\cite{DeCarlo} introduce contour information on a human face to improve tracking stability, while Borshukov \emph{et al.}~\cite{Borshukov} employ manual correction. More recently, Bradley \emph{et al.}~\cite{Bradley} proposed an optimisation method constrained by additional tracking information from multiview video sequences. Beeler~\emph{et al.}~\cite{Beeler} then introduced the concept of anchor frames for human face tracking. In this approach, the sequence is decomposed into several clips based on anchor images which are visually similar to a reference frame. Their optimisation method shortens the tracking distance from reference frames to the target frame to help alleviate errors. However, their approach is domain specific (faces), and assumes that the entire face will return to a neutral expression (the anchor) several times throughout the sequence. In general, it is difficult to label anchor frames on general object sequences with large displacement motion e.g. waving cloth, as there is usually significant deformation between the reference frame and the other frames. In addition, repeated patterns are typically not global as observed in a face (return to a neutral expression). Rather, they occur in smaller local regions at intermittent intervals.

\begin{table*}[t!]
\centerline{
\newcommand{\tabitem}{~~\llap{\textbullet}~~}
\begin{tabular}{lc}
\toprule
\textbf{Input}: A reference frame, a triangle mesh and an image sequence\\
\midrule
\textbf{Step 1.} Computing Optical flow fields (Sec.~\ref{APO:sec:computeOF})\\
\hspace{10 pt}
\textbf{1.1} Compute optical flow fields in both forward ($\textbf{w}_{i \to i+1}$) and backward ($\textbf{w}'_{i+1 \to i}$) \\
\hspace{10 pt}
\textbf{1.2} Define the \emph{Error Score} function\\
\textbf{Step 2.} Detect anchor frames and propagate the entire mesh to these frames (Sec.~\ref{APO:sec:detectAF})\\
\hspace{10 pt}
\textbf{2.1} Match SIFT features from the reference to every other frame\\
\hspace{10 pt}
\textbf{2.2} Compute the general \emph{Error Score} on matchings\\
\hspace{10 pt}
\textbf{2.3} Label the anchor frames from any frame with the low general \emph{Error Score}\\
\textbf{Step 3.} Label anchor patches on non-anchor frames (Sec.~\ref{APO:sec:labelOP})\\
\hspace{10 pt}
\textbf{3.1} Reuse the SIFT feature matching from \textbf{2.1}\\
\hspace{10 pt}
\textbf{3.2} Propagate patches from the reference using \emph{Barycentric Coordinate Mapping}\\
\textbf{Step 4.} Track remaining patches from anchor frames to non-anchor frames (Sec.~\ref{APO:sec:meshTracking})\\
\hspace{10 pt}
\textbf{4.1} Propagate patches from the reference to anchor frames (Sec.~\ref{APO:sec:trackingRtoA})\\
\hspace{25 pt}
\textbf{4.1.1} Compute concatenating optical flow field $\textbf{w}_{R \to A}$\\
\hspace{25 pt}
\textbf{4.1.2} Propagate patches from anchor frames to non-anchor frames using $\textbf{w}_{R \to A}$\\
\hspace{25 pt}
\textbf{4.1.3} Refine the patches using \emph{Error Score}\\
\hspace{10 pt}
\textbf{4.2} Propagate patches from anchor frames to non-anchor frames (Sec.~\ref{APO:sec:trackingAtoN})\\
\hspace{25 pt}
\textbf{4.2.1} Track the patches from the reference to frame $i$ using $\textbf{w}_{A \to i}$\\
\hspace{25 pt}
\textbf{4.2.2} Track the patches from the \emph{Nearest Anchor Patches} to frame $i$\\
\hspace{25 pt}
\textbf{4.2.3} Eliminate the vertex position conflicts between \textbf{4.2.1} and \textbf{4.2.2}\\
\midrule
\textbf{Output}: A mesh tracked throughout the entire image sequence \\
\bottomrule
\end{tabular}
}
\caption{The major steps of the \emph{Anchor Patch} optimisation framework.}
\label{APO:tab:apProcess}
\end{table*}

In this work, we focus on tracking long video sequences using optical flow algorithms, and specifically concentrate on reducing \emph{drift}. The general strategy of our approach is to shorten tracking distances for local regions throughout a long sequence. Our proposed framework combines long term feature matching with optical flow estimation. It may be applied to the tracking of general objects with large displacement motion, and results in a significant reduction in \emph{drift}. We first detect \emph{Anchor Frames} for a sequence (Sec.~\ref{APO:sec:detectAF}). This provides an initial set of start points for tracking the sequence. Our main contribution is extending this approach by proposing the concept of \emph{Anchor Patches} (Sec.~\ref{APO:sec:labelOP}). These are corresponding points and patches throughout the sequence which are propagated directly from the reference frame. Our framework substantially reduces overall drift on a tracked image sequence, and may be applied to any optical flow algorithm in a straightforward manner. In our evaluation, we apply the proposed optimisation framework on 6 popular optical flow estimation algorithms to illustrate it's applicability. We provide analysis of our method using 6 synthetic benchmark sequences (Sec.~\ref{APO:sec:EvaluationAP}) generated using a method similar to~\cite{Garg}, three of which are degraded by adding occlusion, gaussian noise and salt\&pepper noise. In addition, we show its applicability on a popular publicly available real world facial sequence with manually annotated ground truth. We show that our proposed optimisation framework significantly improves tracking accuracy and reduces overall drift when compared against the baseline optical flow approaches alone.

This paper is organized as follows: In Sec.~\ref{APO:sec:overview}, an overview of our proposed optimisation framework is outlined. Sec.~\ref{APO:sec:computeOF}, \ref{APO:sec:detectAF}, \ref{APO:sec:labelOP} and \ref{APO:sec:meshTracking} give details of the four major steps in our framework. In Sec.~\ref{APO:sec:EvaluationAP}, we evaluate our approach using 6 optical flow algorithms tested on 6 synthetic benchmark sequences and a real world facial sequence.

\begin{figure*}[t!]
\begin{center}
\includegraphics[width=0.8\linewidth]{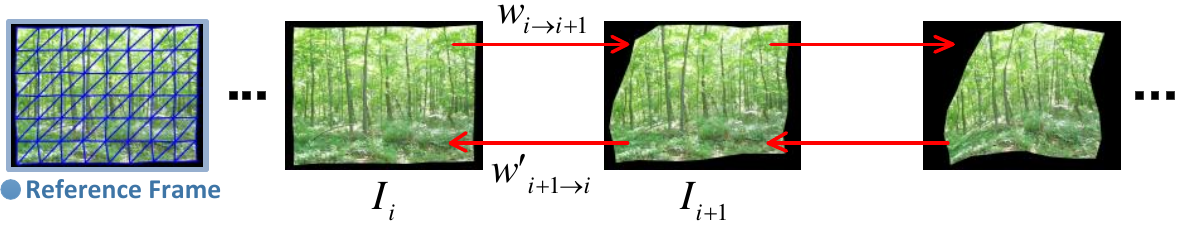}
\end{center}
\vspace{-0.5cm}
\caption{\textbf{Step One.} The optical flow fields are computed in both forward ($\textbf{w}_{i \to i+1}$) and backward ($\textbf{w}'_{i+1 \to i}$) directions between every adjacent images pair in the sequence where the first frame is labelled as a reference frame.}
\label{APO:fig:step_A}
\end{figure*}

\vspace{-0.5cm}

\section{System Overview}
\label{APO:sec:overview}

Our proposed optimisation framework reduces overall optical flow drift given long image sequences, and provides additional robustness against other issues such as large displacements and occlusions. The major procedure is shown in Table~\ref{APO:tab:apProcess}.The aim of our \emph{Anchor Patch optimisation Framework} (APO) is accurately tracking a mesh denoted by $M_{R} = (V_{R},E_{R},F_{R})$ from a reference frame $I_{R}$ to every other frame $I_{i}$ in the sequence. $M_{i} = (V_{i},E_{i},F_{i})$ denotes the corresponding mesh on frame $I_{i}$. In the following sections, the four major steps are discussed in detail.

\vspace{-0.5cm}

\section{Step One: Computing Optical Flow Fields}
\label{APO:sec:computeOF}

The first step is to compute an optical flow field between every frame and its successor over a long video sequence in both forward and backward directions (Fig.~\ref{APO:fig:step_A}). In our evaluation, we consider application of our APO framework on a number of dense correspondence optical flow or tracking approaches, e.g. Brox~\emph{et al.}~\cite{Brox_Large}, Classic+NL~\cite{Sun10} and ITV-L1~\cite{ITV_L1}. Let $\textbf{w}_{i \to i+1}$ denote the optical flow field from frame $I_{i}$ to frame $I_{i+1}$. Similarly we have $\textbf{w}'_{i+1 \to i}$ denoting the optical flow field from frame $I_{i+1}$ to frame $I_{i}$ in the backward direction. The optical flow field between frame $I_{i}$ and $I_{j}$ where $i<j$ (Forward direction), is denoted by $\textbf{w}_{i \to j}$ as $\textbf{w}_{i \to j} = \sum_{i<j}\textbf{w}_{i \to i+1}$. Similarly, the optical flow field between frame $I_{j}$ and $I_{i}$ where $i<j$ (Backward direction), is denoted by $\textbf{w}'_{j \to i}$ as $\textbf{w}'_{j \to i} = \sum_{j>i}\textbf{w}'_{j \to j-1}$.

In order to evaluate the optical flow at a specific pixel $\textbf{x}=(x,y)^T$, an \emph{Error Score} $E(w)$ is proposed here, where $w=(u,v)^T$ is the optical flow vector at pixel $\textbf{x}$. The pixel $\textbf{x}$ in frame $I_{i}$ is matched to pixel $\textbf{x}'=(x',y')^T$ in frame $I_{i+1}$ where $\textbf{x}' = \textbf{x}+w$. The \emph{Error Score} $E(w)$ is calculated as the weighted \emph{Root Mean Square} (RMS) error at a $3 \times 3$ pixel area centred on pixel $\textbf{x}$ and $\textbf{x}'$.

\begin{eqnarray}
E(w)&=&\sqrt{\frac{\alpha_{1}d(x,y)+\alpha_{2}d_{c}(x,y)+\alpha_{3}d_{d}(x,y)}{\alpha_{1}+\alpha_{2}+\alpha_{3}}} \nonumber \\
d_{d}(x,y)&=&d(x-1,y-1)+d(x+1,y+1) \nonumber\\ &+&d(x-1,y+1)+d(x+1,y-1) \nonumber \\
d_{c}(x,y)&=&d(x-1,y)+d(x+1,y) \nonumber\\ &+&d(x,y-1)+d(x,y+1) \nonumber \\
d(x,y) &=& |I_{i}(x,y)-I_{i+1}(x+u,y+v)|^2
\label{APO:eq:errorS}
\end{eqnarray}

Where $\alpha_{1}$, $\alpha_{2}$ and $\alpha_{3}$ are weights for controlling the contribution of each pixel in the $3 \times 3$ area. Using this $3 \times 3$ kernel is supposed to give extra robustness against the subpixel accuracy and illumination changes~\cite{LME,moBlur,li2013nonrigid,moblur_nc}. In our experiments, all these weights are set as $\alpha_{1}=1$, $\alpha_{2}=0.25$ and $\alpha_{3}=0.125$ which refer to the distance from the centre pixel $\textbf{x}$ of the area. This \emph{Error Score} is intended to evaluate the optical flow at a specific pixel. We also use it to evaluate feature matching scores later in our framework.

\vspace{-0.5cm}

\section{Step Two: Labeling Anchor Frames}
\label{APO:sec:detectAF}

\begin{figure*}[htb]
\begin{center}
\includegraphics[width=1\linewidth]{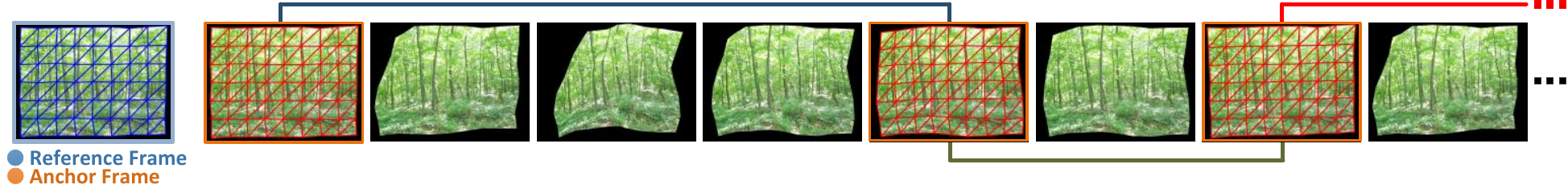}
\end{center}
\vspace{-0.5cm}
\caption{\textbf{Step Two.} The frames are detected as anchor frames (Red) because of the similar appearance to the reference (Blue). These anchor frames partition the entire sequence into several independent clips which allows tracking performing in parallel.}
\label{APO:fig:step_B}
\end{figure*}

After obtaining our optical flow fields, anchor frames are then detected in a similar manner to Beeler~\emph{et al.}~\cite{Beeler}, with the difference that we employ \emph{SIFT} for feature matching as opposed to \emph{Normalised Cross Correlation} (NCC), and additionally use our \emph{Error Score} function (Sec.~\ref{APO:sec:computeOF}) to evaluate matches. The main procedure is as follows (Fig.~\ref{APO:fig:step_B}):

\begin{figure*}[htb]
\begin{center}
\includegraphics[width=1\linewidth]{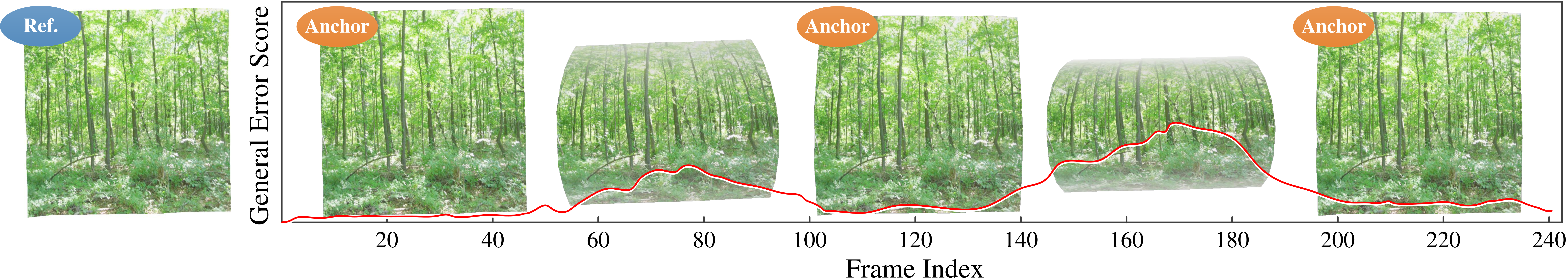}
\end{center}
\vspace{-0.5cm}
\caption{The anchor frames are selected based on our general error score which is computed by comparing the reference frame to every other frame in our \emph{Carton} benchmark sequence.}
\label{APO:fig:anchorSelected}
\end{figure*}

\begin{itemize}
\item \textbf{Feature Capture.} A set of \emph{SIFT} features $S_{R}$ is detected in the reference frame $I_{R}$. Note that other features could be employed, but we select \emph{SIFT} due to the general high accuracy and robustness. Here we apply the GPU version matching approach~\cite{GPU_SIFT} to perform correspondence matching of \emph{SIFT} feature sets $S_{R}$ to feature set $S_{i}$ of any other frames $I_{i}$.
\item \textbf{Outlier Rejection.} The aim of this selection process is removing outliers from our feature matching on all the frames. Correspondence matches of the \emph{SIFT} feature set $S_{R}$ between the reference frame $I_{R}$ and the target frame $I_{i}$ are performed. We select the matches which meet $|\textbf{x}-\textbf{x}'|<\tau$ where $\textbf{x}$ is feature position in $I_{R}$, $\left \{ \textbf{x}\in S_{R},\textbf{x}=(x,y)^T \right \}$; $\textbf{x}'$ is the corresponding feature position in $I_{i}$; $\tau$ is a threshold which is set as 30 pixels in our experiments. We find this simple outlier rejection strategy sufficient for most of cases in our experiments (Sec.~\ref{APO:sec:EvaluationAP}). More sophisticated outlier rejection method such as~\cite{Pizarro} could also be employed.
\item \textbf{General Error Score.} The general error score is computed for every image as the average of the overall \emph{Error Score} $E(w)$ (Eq.~\eqref{APO:eq:errorS}). Frames that contain the lowest general error score (below a specific threshold) are selected as anchor frames denoted $I_{A}$ and the other frames are non-anchor frames. It is because that the general error score is supposed to quantise the general appearance deformation where low score presents the small appearance change. Fig.~\ref{APO:fig:anchorSelected} shows this process on our \emph{Carton} benchmark sequence.
\end{itemize}

After labeling anchor frames that are visually similar to reference frame, these are used as a basis to partition the entire image sequence into several independent \emph{clips}. This also allows computation in the next steps to be performed in parallel. In addition, the mesh $M_{R}$ is propagated from the reference frame $I_{R}$ to each anchor frame $I_{A}$ using \emph{SIFT} matches and a direct optical flow field between them. More detail can be found in Sec.~\ref{APO:sec:trackingRtoA}. The propagated mesh in an anchor frame is denoted $M_{A} = (V_{A},E_{A},F_{A})$. Because of large displacement motion between anchor frames, and the fact that many images in a deformable sequence may not return to a reference point, these alone are typically insufficient to provide reliable tracking. In the next section, the \emph{Anchor Patch} concept will be introduced to overcome this issue.

\section{Step Three: Labeling Anchor Patches}
\label{APO:sec:labelOP}

\begin{figure*}[htb]
\begin{center}
\includegraphics[width=0.8\linewidth]{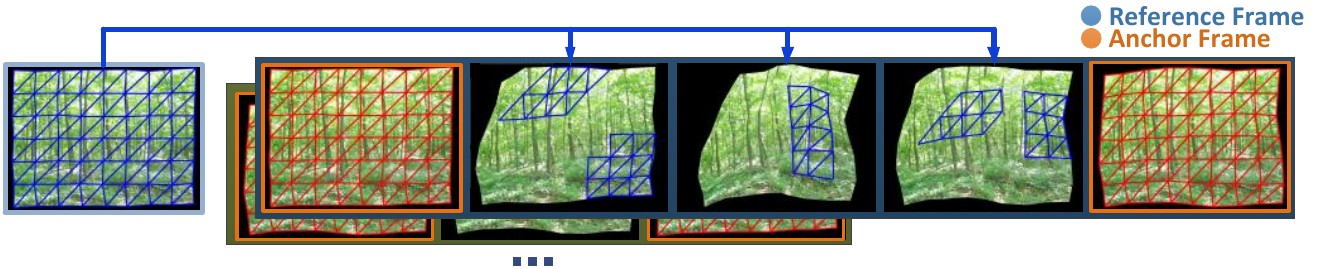}
\end{center}
\vspace{-0.4cm}
\caption{\textbf{Step Three.} Anchor patches (blue patches) are label on non-anchor frames within every clip using \emph{SIFT} feature matching and \emph{Barycentric Coordinate Mapping} between reference frame and non-anchor frame.}
\label{APO:fig:step_C}
\end{figure*}

\begin{figure}[htb]
\begin{center}
\includegraphics[width=1\linewidth]{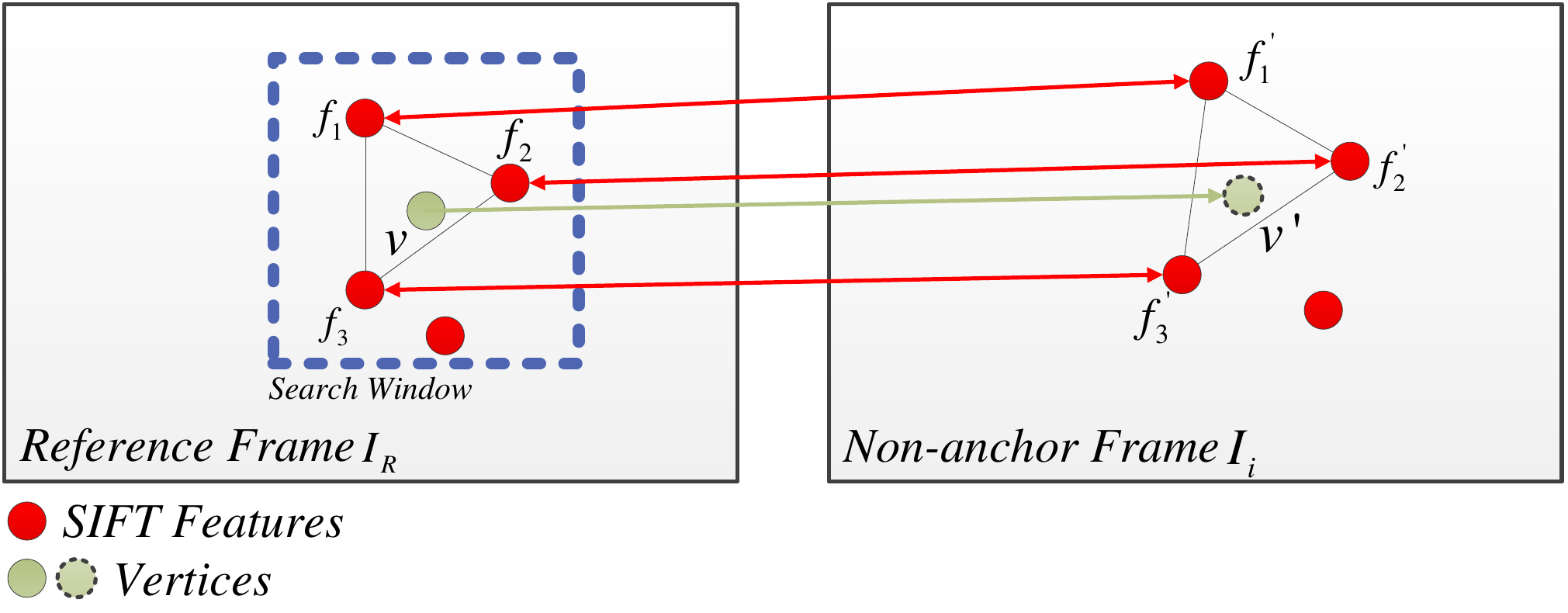}
\end{center}
\vspace{-0.5cm}
\caption{Anchoring patches using \emph{Barycentric Coordinate Mapping} and \emph{SIFT} features.}
\label{APO:fig:triangularLocation}
\end{figure}

The motivation of the original \emph{Anchor Frame} method~\cite{Beeler} is to provide multiple \emph{Starting Points} for tracking. Since error accumulates, the technique is intended to reduce overall error accumulation across long image sequences. However, as mentioned in the previous section, large displacement motion and complex motion may yield a fact that most images in a video sequence have significant visual differences from the reference frame.

The main observation in long image tracking is that local spatial patterns throughout a sequence may be repeated - i.e. part of a cloth might return to the same position several times throughout a video. We take advantage of these repeating regions in order to track between shorter segments, and thus alleviate error accumulation. Apart from taking an entire image as anchor information, an \emph{Anchor Patch} is defined as a set of individual vertices or a group of pixels in the non-reference frame (any other frame in the sequence), which are highly correspondent to a specific part of the reference. The benefit of using anchor patches is to provide additional information for correcting accumulated errors when tracking using optical flow. This technique can also reduce the impact of a low-quality anchor frame (i.e. the one is too dissimilar from the reference frame). Before anchoring patches on non-anchor frames, we first obtain a set of high-quality \emph{SIFT} feature matches between the reference frame and non-anchor frames, i.e. those frames are not already labelled as the reference frame, or an existing anchor frame. This process proceeds as follows:

\begin{itemize}
\item \textbf{Feature Capture.} In order to save the computational time, we reuse the \emph{SIFT} feature sets from \emph{Step Two} (Sec.~\ref{APO:sec:detectAF}). Here the \emph{SIFT} feature set is denoted as $S_{R}$ in the reference frame $I_{R}$; $S_{i}$ presents a feature set of non-anchor frame $I_{i}$.
\item \textbf{Matching Selection.} We also reuse the refined matchings from \emph{Step Two} (Sec.~\ref{APO:sec:detectAF}). This process generates a matches set $\textbf{m}_{R \to i}$ from $S_{R}$ to $S_{i}$.
\end{itemize}

The set of matches $\textbf{m}_{R \to i}$ is used as our initial basis for anchoring patches on non-anchor frames. In order to obtain final anchor patches, \emph{Barycentric Coordinate Mapping} and \emph{Error Refinement} are applied as follows:

\subsubsection{Barycentric Coordinate Mapping}

We suppose to determine the pixel position in a non-anchor frame which corresponds to the position of a vertex on the reference mesh $M_{R}$ in $I_{R}$. These correspondences provide our baseline for stable tracking throughout the image sequence. Fig.~\ref{APO:fig:triangularLocation} illustrates the process of anchoring patches where $v=(x,y)^T$ denotes a vertex in $M_{R}$; $f_{*}=(x_{*},y_{*})^T$, and denotes \emph{SIFT} features in the reference frame $I_{R}$. Similarly, $f'_{*}=(x'_{*},y'_{*})^T$ denotes \emph{SIFT} features in a non-anchor frame $I_{i}$. For the non-anchor frame $I_{i}$, we have $\left \{ f_{k} \to f'_{k} \in \textbf{m}_{R \to i}, k=1,2,3 \dots \right \}$ which denotes previously obtained corresponding \emph{SIFT} feature matches. We wish to calculate the new vertex position $v'=(x',y')^T$ in the non-anchor frame $I_{i}$. We do this by searching for the three nearest \emph{SIFT} features $f_{*}$ in a small $5 \times 5$ search window centred on the vertex of interest $v$. Next, $v'$ is calculated by solving the \emph{Barycentric Coordinate Mapping} equations as:

\begin{eqnarray}
\begin{array}{cc}
\left [
\begin{array}{ccc}
f_1 & f_2 & f_3 \\
f'_1 & f'_2 & f'_3
\end{array}
\right ]
\left [
\begin{array}{c}
\beta_1 \\
\beta_2 \\
\beta_3
\end{array}
\right ]
\end{array}
=
\left [
\begin{array}{c}
v \\
v'
\end{array}
\right ]
\label{APO:eq:triLoc}
\end{eqnarray}

Where $\beta_{*}$ are intermediate variables that satisfy $\beta_{1}+\beta_{2}+\beta_{3}=1$. In practice we found this technique to provide an accurate transformation when applied to small region ($5 \times 5$ pixel block). However, more sophisticated (although slower) interpolation methods could also be used. The process is performed on every vertex in $M_{R}$.

\subsubsection{Error Refinement}

After \emph{Barycentric Coordinate Mapping}, candidate anchor patches denoted by $v'_{*}$ are obtained in non-anchor frames $I_{i}$. We also have matches $v_{*} \to v'_{*}$, the strength of which can be evaluated using our error equation~\eqref{APO:eq:errorS}. Using this error, we select final anchor patches in a non-anchor frame $I_{i}$ using $\left \{P(v'_{*})|E(v_{*} \to v'_{*})< \eta \right \}$ where $\eta$ is a predefined threshold.

\section{Step Four: Mesh Propagation}
\label{APO:sec:meshTracking}

The objective of our optimisation framework is to track a mesh $M_{R}$ from the reference frame to every other frame in an image sequence.
Given tracking information from the previous sections, this process is separated into two steps: first, the mesh $M_{R}$ is propagated from reference frame to anchor frames (Sec.~\ref{APO:sec:detectAF} and \ref{APO:sec:trackingRtoA}). Second, the propagated mesh $M_{A}$ is propagated from anchor frames to the non-anchor frames within the clip (Sec.~\ref{APO:sec:trackingAtoN}).

\begin{figure*}[htb]
\begin{center}
\includegraphics[width=0.8\linewidth]{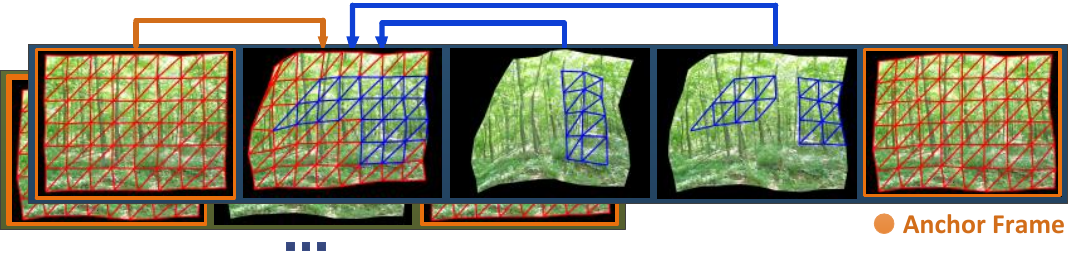}
\end{center}
\vspace{-0.5cm}
\caption{\textbf{Step Four.} Tracking other patches from the anchor frame and nearest anchor patches within a clip where the blue patches are anchor patches, selected from \emph{Nearest Anchor Patch}.}
\label{APO:fig:step_D}
\end{figure*}

\subsection{Propagating from the reference frame to anchor frames}
\label{APO:sec:trackingRtoA}

The mesh propagation process from the reference frame to the anchor frame is as follows:

\begin{itemize}
\item \textbf{Computing the optical flow field.} The optical flow field $\textbf{w}_{R \to A}$ directly between the reference frame to the anchor frame is computed by sum up pairwise optical flow fields $\textbf{w}_{i \to i+1}$ in between.
\item \textbf{Matching selection.} We propagate the whole mesh $M_{R}$ from the reference to the anchor frames. For every vertex in $M_{R}$, \emph{high error matches} in anchor frames are eliminated (see Error Refinement).
\item \textbf{Barycentric Coordinate Mapping.} The positions of those eliminated vertices are recomputed by applying \emph{Barycentric Coordinate Mapping} to low error matches. The operation is shown in Fig.~\ref{APO:fig:triangularLocation}.
\end{itemize}

After this stage, information for every vertex in $M_{R}$ is established from the reference frame to the anchor frame.

\subsection{Propagating from anchor frames to non-anchor frames}
\label{APO:sec:trackingAtoN}

The entire image sequence is partitioned into clips which are bound by different anchor frames. The propagation process can be individually performed within these clips in parallel. Within these clips, the anchor patches are supposed to improve overall tracking stability and accuracy. In order to use anchor patches in this process, we define \emph{Nearest Anchor Patch} as follows. For vertex $v$ in $M_{A}$, the \emph{Nearest Anchor Patch} of $v$ on frame $I_{i}$ is the anchor patch $\left \{ v'_{i+k}|v \to v'_{i+k} \right \}$ on non-anchor frame $I_{i+k}$ which is nearest to $I_{i}$ in the image sequence. Fig.~\ref{APO:fig:trackAtoNimg} shows an example where frame $I_{i+k}$ is the frame which is nearest to frame $I_{i}$ in image sequence and contains anchor patch $v'_{i+k}$ matching to $v$ in anchor frame $I_{A}$. The main tracking procedure proceeds (Fig.~\ref{APO:fig:step_D}) as follows:

\begin{figure*}[htb]
\begin{center}
\includegraphics[width=0.8\linewidth]{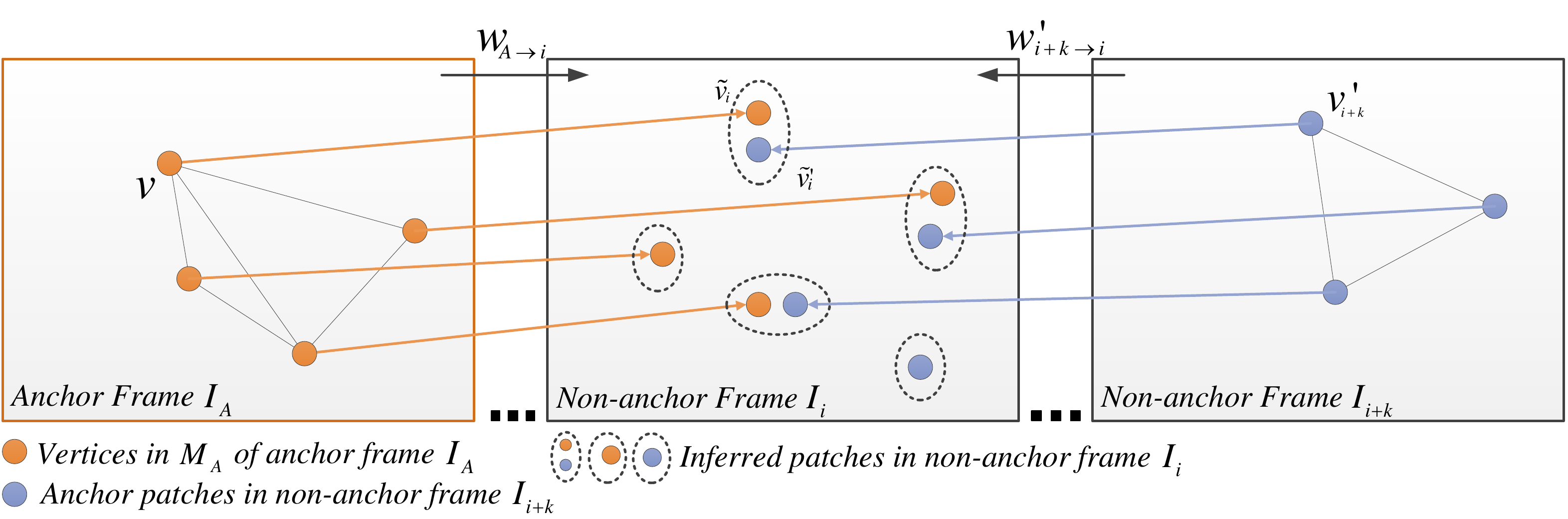}
\end{center}
\vspace{-0.5cm}
\caption{Vertex conflict can happen when mesh and anchor patches are propagated to target frame $I_{i}$. Here $v'_{i+k}$ is an anchor patch that is strongly matched to $v$.}
\label{APO:fig:trackAtoNimg}
\end{figure*}

\begin{itemize}
\item \textbf{Mesh propagation.} In order to establish tracking information between anchor frames and non-anchor frame, the mesh $M_{A}$ is first propagated from anchor frame $I_{A}$ to non-anchor frames $I_{i}$ using the previously calculated optical flow field $\textbf{w}_{A \to i}$ from \emph{Step One} (Sec.~\ref{APO:sec:computeOF}).
\item \textbf{Anchor patches propagation.} The \emph{Nearest Anchor Patch} of each vertex $v$ in $M_{A}$ is searched through the whole clip then propagated to non-anchor frame $I_{i}$ using the optical flow field in the forward $\textbf{w}_{* \to i}$ or backward $\textbf{w}'_{i+k \to i}$ direction.
\item \textbf{Conflict eliminating.} After propagating the mesh and nearest anchor patches to non-anchor frame $I_{i}$, there may be position conflict on some of the propagated vertices. As shown in Fig.~\ref{APO:fig:trackAtoNimg}, $\tilde v_i$ and $\tilde v'_i$ are not in the same desired position. In order to eliminate the conflict, the position of $\left \{v_{i}|v \to v_{i} \right \}$ matching to $v$ can be calculated using the sum of all weighted candidate positions e.g. $\tilde v_i$ and $\tilde v'_i$ (Eq.\ref{APO:eq:conf}) based on the \emph{Error Score}.
    \begin{equation}
    v_i = \frac{E(v \to \tilde v'_i) \tilde v_i + E(v \to \tilde v_i)\tilde v'_i}{E(v \to \tilde v'_i)+E(v \to \tilde v_i)}
    \label{APO:eq:conf}
    \end{equation}
\end{itemize}

Due to the fact that the anchor frames divide the overall sequence into smaller clips, this allows the mesh propagation in between to be calculated in parallel. In the next section we perform an evaluation of our framework.

\section{Evaluation}
\label{APO:sec:EvaluationAP}

We evaluate APO with a range of 6 popular optical flow estimation methods which are publicly available from the \emph{Middlebury Evaluation System}~\cite{Middlebury}. \emph{Combined local-global Optical Flow} (CLG-TV)~\cite{CLG-TV1}, \emph{Large Displacement Optical Flow} (LDOF)~\cite{Brox_Large} and Classic+NL~\cite{Sun10} are state of the art while the \emph{Horn and Schunck} (HS)~\cite{HS}, \emph{Black and Anandan} (BA)~\cite{BA,Sun10}, \emph{Improved TV-L1} (ITV-L1)~\cite{ITV_L1} are classic optical flow frameworks and also widely used. CLG-TV is a high speed approach that uses a combination of bilateral filtering and anisotropic regularization and also one of the top three algorithms in the normalized interpolation error test from Middlebury. LDOF is an integration of rich feature descriptors and variational optical flow and one of best current optical flow estimation algorithms for large displacement motion. Classic+NL provides high performance in the Middleburry evaluation by formalizing the median filtering heuristic and Lorentzian penalty as explicit objective functions in an \emph{improved TV-L1} framework. The HS method is a pioneering technique optical flow. BA provides improvements to the HS framework by introducing robust quadratic error formulation. ITV-L1 is a recent and increasingly popular optical flow framework which uses a similar numerical optimisation scheme to Classic+NL. Our choice of a mixture of newer, state of the art methods, with older traditional approaches, is to highlight the fact that irrespective of the approach used, our APO framework provides significantly improved tracking in all cases.

\begin{table*}[t!]
 \centerline{
    {\scriptsize
    \begin{tabular}{|c||cccc|ccc|}
    \hline
        & \multicolumn{7}{c|}{Information of the Benchmark Sequences}\\
        & \multicolumn{1}{c}{Original}
        & \multicolumn{1}{c}{Occlusion}
        & \multicolumn{1}{c}{Guass.N}
        & \multicolumn{1}{c|}{S\&P.N}
        & \multicolumn{1}{c}{Carton}
        & \multicolumn{1}{c}{Serviette}
        & \multicolumn{1}{c|}{Frank}\\
        \hline
        \textbf{Image Size (pix.)}             & $500\times500$  & $500\times500$ & $500\times500$ & $500\times500$ & $1024\times768$ & $1024\times768$ & $720\times576$\\
        \textbf{Sequence Length}        & 237 & 237 & 237 & 237 & 266 & 307 & 300\\
        \textbf{Annotation Points}      & 160 & 160 & 160 & 160 & 81 & 63 & 68\\
        \textbf{Avg. Feature Amount}    & 364.80 & 358.32 & 566.13 & 1276.50 & 2498.01 & 3315.49 & 2071.11\\
        \hline
    \end{tabular}
    }
    }
    \caption{An overview of the benchmark sequences in our evaluation. That includes 4 attributes of image size (pixel), sequence length, number of ground truth annotation points per frame and average SIFT feature amount per frame.}\vspace{-3mm}
    \label{APO:tab:infoSeqs}
\end{table*}

\begin{figure}[t!]
\begin{center}
\includegraphics[width=0.95\linewidth]{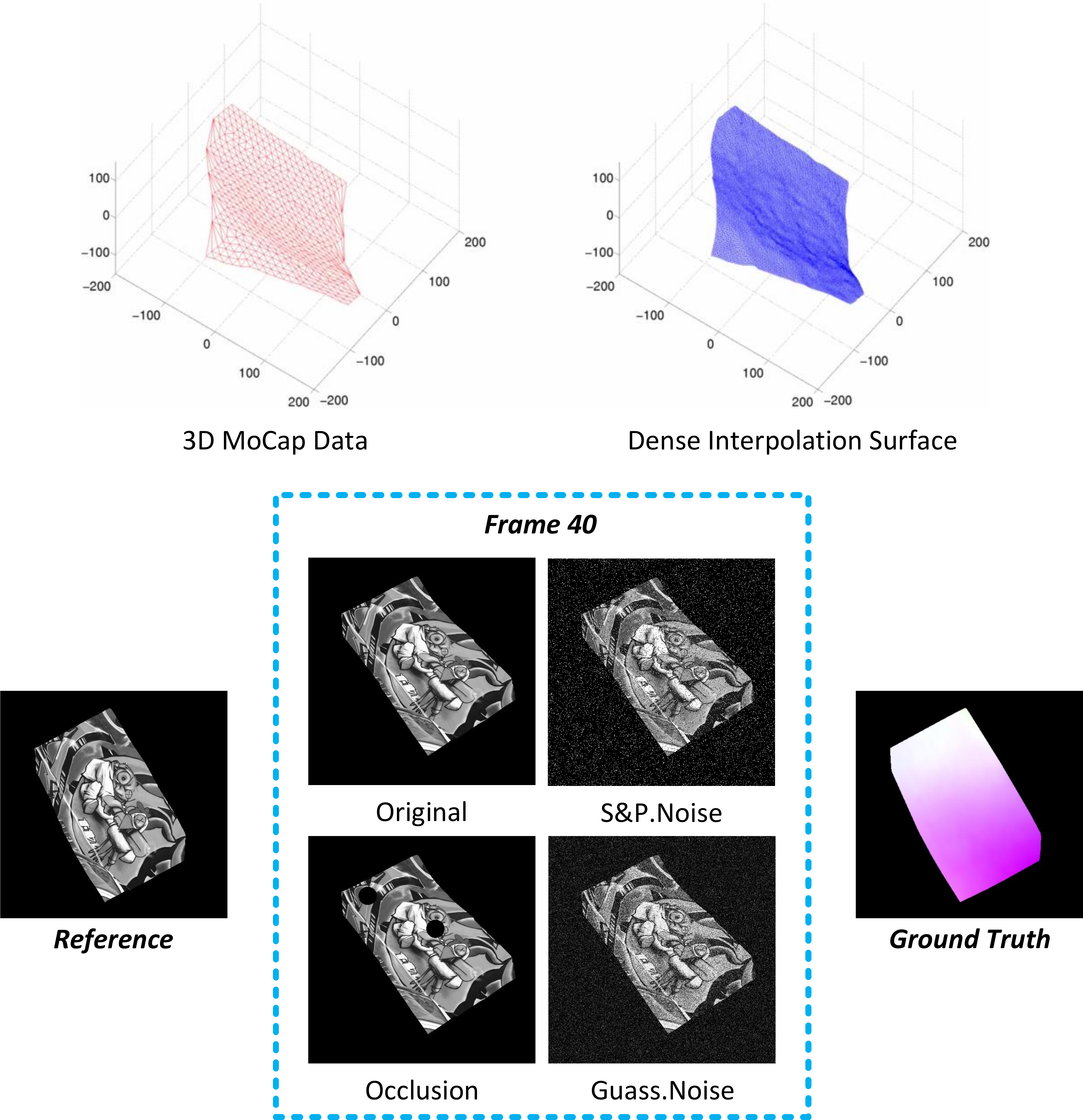}
\end{center}
\vspace{-0.5cm}
\caption{Our long nonrigid synthetic sequences with ground truth are rendered by using flag MOCAP data~\cite{White} and Garg~\emph{et al.}~\cite{Garg13}. Images are adopted from~\cite{Garg}}\vspace{-1mm}
\label{APO:fig:queenmary}
\end{figure}

For our evaluation, we compare the optical flow estimation methods previously mentioned -- with and without our optimisation framework -- on 7 long benchmark sequences with ground truth. Table~\ref{APO:tab:infoSeqs} gives an overview of the benchmark sequences used in our evaluation. In previous work Garg~\emph{et al.} released to the community a set of ground truth data for evaluating optical flow algorithms over long sequences. This is as opposed to the Middlebury dataset, which just considers optical flow between pairs of images, and is therefore not applicable to our framework. The sequences of Garg~\emph{et al.} contains 60 frames and are generated using interpolated dense \emph{Motion Capture} (MOCAP) data from real deformations of a waving flag~\cite{White}. As shown in Fig.~\ref{APO:fig:queenmary}, we use the same MOCAP data to generate a long video sequence and three other degraded sequences, each of which contains 237 frames of size $500 \times 500$ pixels. The three degraded sequences are generated in order to test the robustness of our APO framework under different image conditions. They are generated by individually adding synthetic occlusions, gaussian noise and salt \& pepper noise with the same parameters described in~\cite{Garg}. In order to increase the diversity of the sequences, we include three other sequences. One is a \emph{Talking Face Video} (Frank) sequence which contains 300 frames with 68 ground truth annotation points per frame. The other two are also synthetic benchmark sequences generated using MOCAP data of Salzmann \emph{et al.}~\cite{Salzmann} from the carton and serviette deformations. One contains 266 frames of size $1024 \times 768$ while the other contains 307 frames of the same image size. In addition, we also consider the effect of the number of SIFT features detected in the frame, and how this affects overall tracking stability of the APO framework. All optical flow algorithms are applied with default parameter settings from their original papers.

Our baseline optical flow based tracking strategy -- for each of the above algorithms -- is performed as follows: First, the optical flow field is computed (in forward direction) for every pair of adjacent frames in the sequence. We then mark the initial tracking points in the first frame using the same ground truth points in the same frame of the sequence (Table~\ref{APO:tab:infoSeqs}). The correspondent points in the next frame are computed based on the optical flow field in between. This process is repeated until correspondent landmark points are obtained in every frame of the sequence. The average \emph{Endpoint Error} (EE)~\cite{Middlebury} is then calculated against the ground truth annotation points. We then apply our APO framework using the same optical flow fields.). Note that the parameter values relevant to the APO framework are initially and experimentally selected, but then remain constant in all our evaluations.

Table~\ref{APO:tab:avgRMS}(a) shows the measurement of average \emph{Endpoint Error} (AEE) in pixels over all the frames of the sequences.  We highlight the top three best AEE measures for each sequence using superscripts next to different values. Notice that APO significantly reduces the AEE compared to the baseline optical flow methods. Our optimisation framework yields the best AEE measure in all the cases. For instance, \emph{ITV-L1} with APO performs the best in sequence \emph{Original} while \emph{LDOF} with APO yields the best result in sequence \emph{Frank}. We also observe that although in the \emph{Guass.Noise} and \emph{S\&P.Noise} sequences the improvement is less than in the unaltered sequences, the overall result is still an improvement with the addition of APO. We also observe that \emph{LDOF} gives good results even without APO. It is because that the \emph{LDOF} framework takes into account both regular optical flow energy and the feature technique. The latter contributes additional accuracy.

\begin{table*}[t!]
 \centerline{
 \subfigure[Average \emph{Endpoint Error} (AEE) comparison of different methods with our optimisation framework on the benchmark sequences.]{
    \begin{tabular}{|c||cccc|ccc|}
    \hline
        & \multicolumn{7}{c|}{Average \emph{Endpoint Error} in pix (AEE)}\\
        \bfseries{Methods}
        & \multicolumn{1}{c}{Original}
        & \multicolumn{1}{c}{Occlusion}
        & \multicolumn{1}{c}{Guass.N}
        & \multicolumn{1}{c|}{S\&P.N}
        & \multicolumn{1}{c}{Carton}
        & \multicolumn{1}{c}{Serviette}
        & \multicolumn{1}{c|}{Frank}\\ 
        \hline
        \textbf{BA}~\cite{BA}           & 6.14 & 8.03 & 11.02 & 7.79 & 10.56 & 5.18 & 17.57\\
        \textbf{BA +~\emph{APO}}        & $1.72^{2}$ & $1.91^{2}$ & $\textbf{7.89}^{1}$ & $\textbf{5.04}^{1}$ & 2.77 & $\textbf{1.56}^{1}$ & 6.60\\
        \textbf{CLG-TV}~\cite{CLG-TV1}  & 8.59 & 10.93 & 20.28 & 33.93 & 28.94 & 32.17 & 19.29\\
        \textbf{CLG-TV +~\emph{APO}}    & 2.25 & 2.97 & 12.31 & 18.99 & 6.95 & 9.43 & 7.05\\
        \textbf{HS}~\cite{HS}           & 29.16 & 30.44 & 29.74 & 29.43 & 27.69 & 37.90 & 31.27\\
        \textbf{HS +~\emph{APO}}        & 11.68 & 12.88 & 17.79 & 17.21 & 10.25 & 10.03 & 14.19\\
        \textbf{LDOF}~\cite{Brox_Large}       & 6.21 & 6.39 & 16.24 & 24.14 & 6.33 & 5.51 & 14.73\\
        \textbf{LDOF +~\emph{APO}}      & $1.75^{3}$ & $\textbf{1.67}^{1}$ & 11.65 & 13.12 & $\textbf{1.18}^{1}$ & $1.84^{2}$ & $\textbf{3.12}^{1}$\\
        \textbf{Classic+NL}~\cite{Sun10}& 7.07 & 10.61 & 12.65 & 9.50 & 5.72 & 6.62 & 17.32\\
        \textbf{Classic+NL +~\emph{APO}}& 2.15 & 3.18 & $8.31^{2}$ & $6.46^{2}$ & $1.34^{2}$ & $2.03^{3}$ & $3.44^{2}$\\
        \textbf{ITV-L1}~\cite{ITV_L1}     & 5.73 & 8.25 & 17.29 & 14.49 & 5.34 & 7.11 & 17.91\\
        \textbf{ITV-L1 +~\emph{APO}}     & $\textbf{1.50}^{1}$ & $2.33^{3}$ & $9.53^{3}$ & $7.70^{3}$ & $1.70^{3}$ & 2.36 & $3.69^{3}$\\
        \hline
    \end{tabular}
    \label{APO:tab:avgRMS_1}
    }
    }

 \centerline{
 \subfigure[Average \emph{Endpoint Error} (AEE) comparison of Garg~\emph{et al.} and Pizarro~\emph{et al.} on the benchmark sequences (directly tracking from the reference to any other frames). \label{APO:tab:avgRMS_2}]{
    \begin{tabular}{|c||cccc|ccc|}
    \hline
        & \multicolumn{7}{c|}{Average \emph{Endpoint Error} in pix (AEE)}\\
        \bfseries{Methods}
        & \multicolumn{1}{c}{Original}
        & \multicolumn{1}{c}{Occlusion}
        & \multicolumn{1}{c}{Guass.N}
        & \multicolumn{1}{c|}{S\&P.N}
        & \multicolumn{1}{c}{Carton}
        & \multicolumn{1}{c}{Serviette}
        & \multicolumn{1}{c|}{Frank}\\ 
        \hline
        Garg~\emph{et al.}, PCA~\cite{Garg} & 0.61 & 0.71 & 1.64 & 1.21 & N/A & N/A & N/A\\
        Garg~\emph{et al.}, PCA~\cite{Garg} +~\textbf{\emph{APO}} & 0.60 & $\textbf{0.69}^{1}$ & 1.65 & 1.23 & N/A & N/A & N/A\\
        Garg~\emph{et al.}, DCT~\cite{Garg} & $\textbf{0.59}^{1}$ & 0.74 & 1.86 & 1.54 & N/A & N/A & N/A\\
        Garg~\emph{et al.}, DCT~\cite{Garg} +~\textbf{\emph{APO}} & $\textbf{0.59}^{1}$ & 0.73 & 1.85 & 1.53 & N/A & N/A & N/A\\
        Pizarro~\emph{et al.}~\cite{Pizarro} & 0.71 & 0.79 & 0.99 & 0.98 & N/A & N/A & N/A\\
        Pizarro~\emph{et al.}~\cite{Pizarro} +~\textbf{\emph{APO}} & 0.79 & 0.81 & $\textbf{0.99}^{1}$ & $\textbf{0.98}^{1}$ & N/A & N/A & N/A\\
        \hline
    \end{tabular}
    }
    }
    \caption{Average \emph{Endpoint Error} (AEE) comparison on our long benchmark sequences.}
    \label{APO:tab:avgRMS}
 \end{table*}

Table~\ref{APO:tab:avgRMS}(b) shows another experiment, in which we performs Garg~\emph{et al.}~\cite{Garg} and Pizarro~\emph{et al.}~\cite{Pizarro} on our benchmark sequences (results on \emph{Carton}, \emph{Serviette} and \emph{Frank} are not available.) using a direct tracking strategy. Here we compute the optical flow fields directly from the reference to any other frames of the sequence. The annotation points are then directly tracked to the test frames using those flow fields. Note that the numbers in Table~\ref{APO:tab:avgRMS}(b) may be slightly different from their original work~\cite{Garg13}. It is because that, first, our sequences are extended to 237 frames which is around 3 times longer; second, we evaluate the tracking results of only 160 annotation points instead of all the pixels. We observe that the both state-of-the-art approaches (Garg~\emph{et al.} and Pizarro~\emph{et al.}) give higher accuracy than any other baseline methods in Table~\ref{APO:tab:avgRMS}(a). The hidden conditions are (1) the tracking distance is minimum for Garg~\emph{et al.} and Pizarro~\emph{et al.} which very much reduces the accumulate errors; (2) both Garg~\emph{et al.} and Pizarro~\emph{et al.} shows high accuracy for nonrigid surface tracking in the record~\cite{Garg,Pizarro}. And all our sequences contain single nonrigid object. However, such direct tracking strategy cannot handle the situation where objects may be temporally out of the scene. In addition, the object appearance in the reference may be significantly different from the one in some other frames of the sequence. That brings extra difficulty to optical flow estimation. In this measure (Table~\ref{APO:tab:avgRMS}(b)), our \textbf{APO} degrades to correct the tracking using the long term features. In the \emph{Orignal} and \emph{Occlusion} cases, the result shows that using \textbf{APO} still yields lower (or the same) errors over the baseline methods even those have given really minuscule measures ( $<$ 1.60 pixel ). We also observe that \textbf{APO} slightly reduces the accuracy on the noisy cases because the extra noises affect the feature detection. Please note that the results are shown as ''N/A`` on \emph{Carton}, \emph{Serviette} and \emph{Frank} as the baselines Garg~\emph{at al.} and Pizarro~\emph{at al.} are not available.

While we consider ourselves primarily with tracking over long sequences, the shorter sequences are consider as well. In Table~\ref{APO:tab:avg30RMS}, the AEE measures of various methods are compared on the first 30 frames of our benchmark sequences. We observe similar AEE measures as in the long sequence case (Table~\ref{APO:tab:avgRMS_1}). The APO framework significantly increases the tracking accuracy -- outperforming the baseline tracking methods in all cases even given degradation (e.g. \emph{Gauss.Noise} and \emph{S\&P.Noise}). Moreover, the \emph{BA} with APO is also observed to overfit in the noisy sequences while \emph{Classic+NL} with APO yields the best measures in both sequences of \emph{Gauss.Noise} and \emph{S\&P.Noise}.

\begin{table*}[t!]
 \centerline{
    \begin{tabular}{|c||cccc|ccc|}
    \hline
        & \multicolumn{7}{c|}{Average \emph{Endpoint Error} (AEE) on the First 30 Frames}\\
        \bfseries{Methods}
        & \multicolumn{1}{c}{Original}
        & \multicolumn{1}{c}{Occlusion}
        & \multicolumn{1}{c}{Guass.N}
        & \multicolumn{1}{c|}{S\&P.N}
        & \multicolumn{1}{c}{Carton}
        & \multicolumn{1}{c}{Serviette}
        & \multicolumn{1}{c|}{Frank}\\ 
        \hline
        \textbf{BA}~\cite{BA}           & 1.57&1.72&3.87&2.71&2.37&$1.56^3$&8.76\\
        \textbf{BA +~\emph{APO}}        & $1.41^3$&1.65&$3.66^3$&$2.13^2$&2.17&$\textbf{1.13}^1$&5.40\\
        \textbf{CLG-TV}~\cite{CLG-TV1}  & 2.40&2.60&6.71&8.77&8.10&5.54&8.60\\
        \textbf{CLG-TV +~\emph{APO}}    & 2.10&2.24&6.53&8.39&4.79&5.11&7.35\\
        \textbf{HS}~\cite{HS}           & 33.67&35.70&35.05&34.50&26.16&22.08&12.76\\
        \textbf{HS +~\emph{APO}}        & 16.11&16.32&13.78&19.37&9.78&6.33&9.19\\
        \textbf{LDOF}~\cite{Brox_Large}       & 2.38&2.37&3.96&4.03&3.90&2.52&8.51\\
        \textbf{LDOF +~\emph{APO}}      & $1.15^2$&$\textbf{0.97}^1$&3.75&2.66&$\textbf{0.89}^1$&$1.44^2$&$\textbf{2.82}^1$\\
        \textbf{Classic+NL}~\cite{Sun10}& 1.63&1.76&$3.61^2$&$2.51^3$&2.18&1.75&8.77\\
        \textbf{Classic+NL +~\emph{APO}}& 1.51&$1.33^2$&$\textbf{3.54}^1$&$\textbf{1.99}^1$&$1.24^2$&1.68&$3.70^3$\\
        \textbf{ITV-L1}~\cite{ITV_L1}     & 1.55&1.76&6.27&5.07&2.37&2.01&9.22\\
        \textbf{ITV-L1 +~\emph{APO}}     & $\textbf{0.99}^1$&$1.31^2$&5.77&4.65&$1.69^3$&1.71&$3.48^2$\\
        \hline
    \end{tabular}
    }
    \caption{Average \emph{Endpoint Error} (AEE) comparison of different methods with our optimisation framework on the first 30 frames of the benchmark sequences.}
    \label{APO:tab:avg30RMS}
 \end{table*}

\begin{table*}[t!]
\centerline{
    \begin{tabular}{|c||cccc|ccc|}
    \hline
        & \multicolumn{7}{c|}{Average \emph{Endpoint Error} (AEE) on Different Feature Distributions}\\
        \bfseries{Methods}
        & \multicolumn{1}{c}{Original}
        & \multicolumn{1}{c}{Occlusion}
        & \multicolumn{1}{c}{Guass.N}
        & \multicolumn{1}{c|}{S\&P.N}
        & \multicolumn{1}{c}{Carton}
        & \multicolumn{1}{c}{Serviette}
        & \multicolumn{1}{c|}{Frank}\\ 
        \hline
        \textbf{BA}~\cite{BA}, \textbf{No \emph{APO}}   & 6.14 & 8.03 & 11.02 & 7.79 & 10.56 & 5.18 & 17.57\\
        \textbf{\emph{APO}, 100\% Feature}        & $1.72^{2}$ & $1.91^{2}$ & $\textbf{7.89}^{1}$ & $\textbf{5.04}^{1}$ & 2.77 & $\textbf{1.56}^{1}$ & 6.60\\
        \textbf{\emph{APO}, 50\% Feature}        & 3.64&4.71&8.06&6.12&5.89&2.98&10.63\\
        \textbf{\emph{APO}, 0\% Feature}        & 5.12&6.44&9.23&7.21&8.69&4.35&12.69\\
        \hline
        \textbf{CLG-TV}~\cite{CLG-TV1}, \textbf{No \emph{APO}}  & 8.59 & 10.93 & 20.28 & 33.93 & 28.94 & 32.17 & 19.29\\
        \textbf{\emph{APO}, 100\% Feature}    & 2.25 & 2.97 & 12.31 & 18.99 & 6.95 & 9.43 & 7.05\\
        \textbf{\emph{APO}, 50\% Feature}    & 4.86&6.51&14.39&22.72&15.36&19.91&12.00\\
        \textbf{\emph{APO}, 0\% Feature}    & 6.94&9.11&16.83&26.03&23.57&24.03&15.07\\
        \hline
        \textbf{HS}~\cite{HS}, \textbf{No \emph{APO}}           & 29.16 & 30.44 & 29.74 & 29.43 & 27.69 & 37.90 & 31.27\\
        \textbf{\emph{APO}, 100\% Feature}        & 11.68 & 12.88 & 17.79 & 17.21 & 10.25 & 10.03 & 14.19\\
        \textbf{\emph{APO}, 50\% Feature}        & 18.13&20.28&20.66&19.91&17.39&25.99&23.45\\
        \textbf{\emph{APO}, 0\% Feature}        & 24.73&27.11&23.97&23.40&24.09&33.11&29.17\\
        \hline
        \textbf{LDOF}~\cite{Brox_Large}, \textbf{No \emph{APO}}       & 6.21 & 6.39 & 16.24 & 24.14 & 6.33 & 5.51 & 14.73\\
        \textbf{\emph{APO}, 100\% Feature}      & $1.75^{3}$ & $\textbf{1.67}^{1}$ & 11.65 & 13.12 & $\textbf{1.18}^{1}$ & $1.84^{2}$ & $\textbf{3.12}^{1}$\\
        \textbf{\emph{APO}, 50\% Feature}      & 3.21&3.09&12.18&15.02&2.90&3.74&8.66\\
        \textbf{\emph{APO}, 0\% Feature}      & 5.08&5.24&14.11&18.46&5.45&4.89&11.76\\
        \hline
        \textbf{Classic+NL}~\cite{Sun10}, \textbf{No \emph{APO}} & 7.07 & 10.61 & 12.65 & 9.50 & 5.72 & 6.62 & 17.32\\
        \textbf{\emph{APO}, 100\% Feature} & 2.15 & 3.18 & $8.31^{2}$ & $6.46^{2}$ & $1.34^{2}$ & $2.03^{3}$ & $3.44^{2}$\\
        \textbf{\emph{APO}, 50\% Feature} & 4.00&6.39&9.48&7.33&3.89&4.00&10.14\\
        \textbf{\emph{APO}, 0\% Feature} & 5.96&7.78&11.64&8.98&4.78&6.00&13.27\\
        \hline
        \textbf{ITV-L1}~\cite{ITV_L1}, \textbf{No \emph{APO}}     & 5.73 & 8.25 & 17.29 & 14.49 & 5.34 & 7.11 & 17.91\\
        \textbf{\emph{APO}, 100\% Feature}     & $\textbf{1.50}^{1}$ & $2.33^{3}$ & $9.53^{3}$ & $7.70^{3}$ & $1.70^{3}$ & 2.36 & $3.69^{3}$\\
        \textbf{\emph{APO}, 50\% Feature}     & 3.59&5.17&10.93&8.47&3.41&5.00&10.11\\
        \textbf{\emph{APO}, 0\% Feature}     & 4.77&6.92&12.50&10.31&4.43&5.95&14.29\\
        \hline
    \end{tabular}
    }
    \caption{Average \emph{Endpoint Error} (AEE) comparison on the benchmark sequences with varying feature distributions.}
    \label{APO:tab:percSiftRMS}
 \end{table*}

We also evaluate the effect on tracking accuracy by varying the number of selected features. Different numbers (50\% and 0\%) of features are randomly selected from the initial full detection feature set before performing \emph{Anchor Patch} detection. Information on our total number of features can be found in Table~\ref{APO:tab:infoSeqs}, e.g. there are 364.80 features averagely on each frame of the sequence \emph{Original}. Table~\ref{APO:tab:percSiftRMS} shows an AEE comparison given various numbers of features. We observe that AEE is improved given more features in all cases. Another interesting observation is that our optimisation framework provides lower error against the baseline tracking strategy even given sparse or no features (0\% feature). Note that in this case, our APO framework defaults to using an optical flow method with just the \emph{Anchor Frame} approach~\cite{Beeler}. Also note -- for example by comparing to Table 3 -- that this indicates that the APO framework also provides significant tracking improvement over using anchor frames alone.

\begin{figure*}[t!]
    \subfigure[Visual comparison of different methods on the frame 88 of the sequence \emph{Frank}.]{\includegraphics[width=0.97\linewidth]{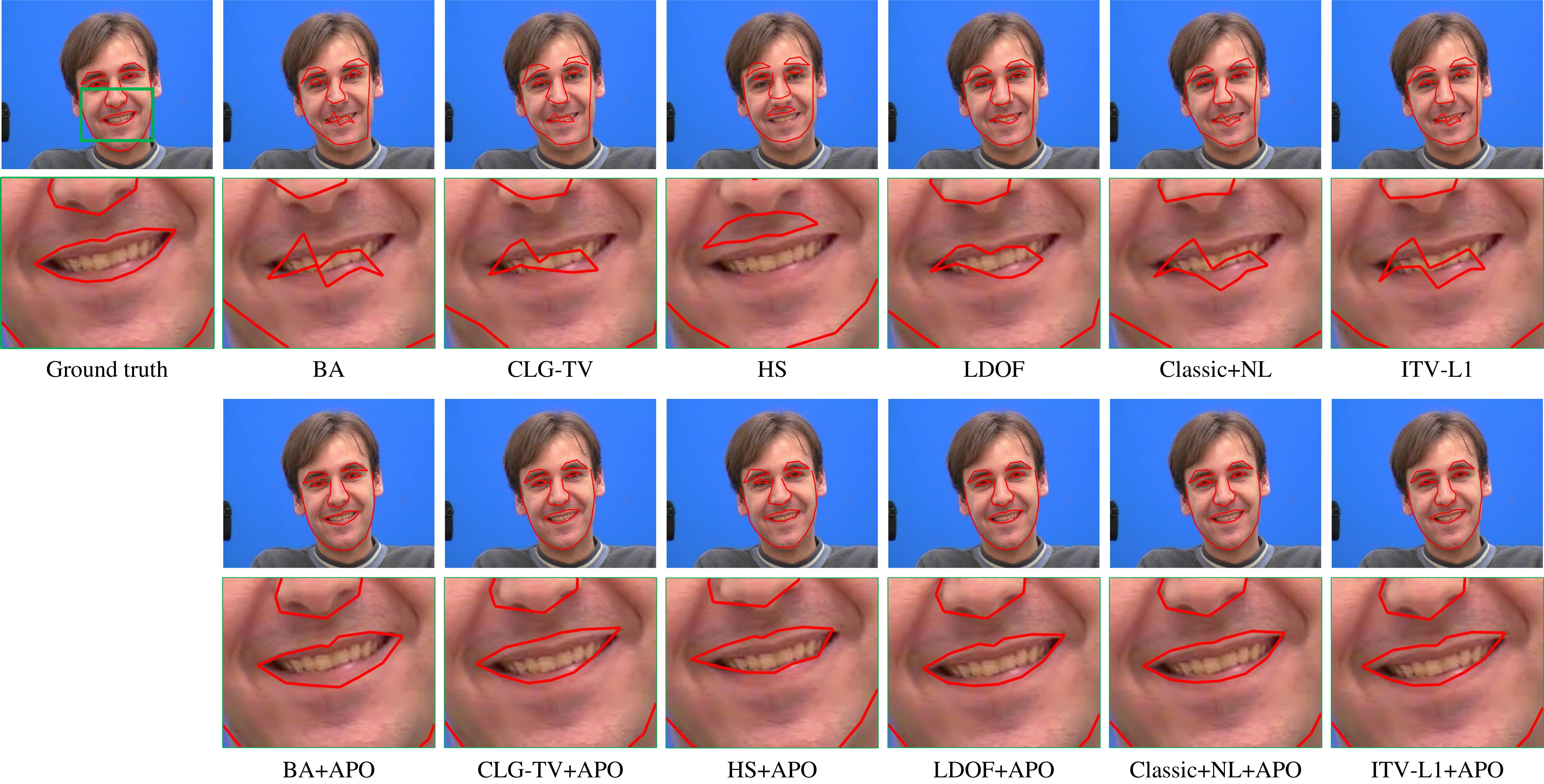}
    \label{APO:fig:visFigsFrank}}\\

    \centerline{
    \subfigure[Visual comparison of different methods on the frame 192 of the sequence \emph{Serviette}.]{\includegraphics[width=0.97\linewidth]{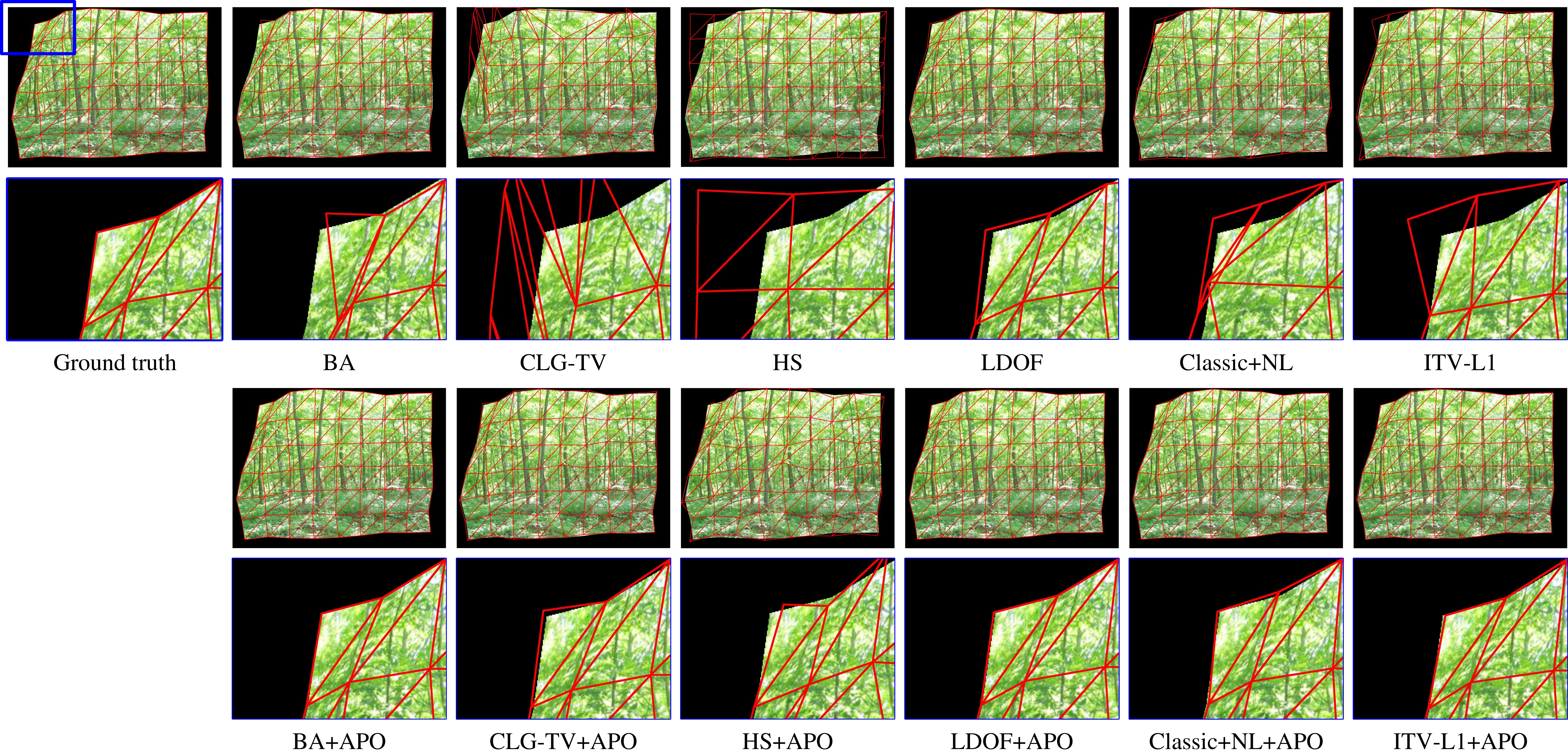}
    \label{APO:fig:visFigsVicon}}
    }\vspace{-2mm}
    \caption{Visual comparison and AEE measures on sequences of \emph{Frank} and \emph{Serviette}.}\vspace{-2mm}
    \label{APO:fig:visFigs}
\end{figure*}

We also make the visual comparisons on two of our sequences, \emph{Frank} and \emph{Serviette}. The former is real world sequence with ground truth annotation points, while the latter is synthetic sequence overlaid with a ground truth mesh. In Fig.~\ref{APO:fig:visFigs}, we observe noticeable \emph{drift} problems given the baseline optical flow tracking strategy. Also note that more details can be found in the corresponding video footage where we visually show that our framework significantly reduces the \emph{drift}.

The computational consumption of our framework heavily relies on the supplementary optical flow method, because we need to calculate the optical flow fields twice (forward and backward) for every pair of adjacent images. Apart from this, our framework can be implemented in a parallel computation fashion. Anchor frames divide the sequence into clips which give multiple start points for tracking. In the implementation, a GPU version of SIFT approach~\cite{GPU_SIFT} is applied for feature detection and matching (around 10 frames per second on our benchmarks). The whole framework is constructed under CUDA platform. Assuming all optical flow fields are obtained, our framework reach real-time efficiency (around 2 frames per second) on our benchmarks using on a 2.9Ghz Xeon 8-cores, NVIDIA Quadro FX 580, 16Gb memory computer.

\vspace{-4mm}
\section{Conclusion}
\label{APO:sec:APO_con}
\vspace{-2mm}

In this paper, we have presented a novel optimisation framework using \emph{Anchor Patches} constraint, which improves the tracking on mesh or sparse points through long image sequences. Our optimisation framework temporally anchors the image regions throughout the sequence in order to mitigate the effect of \emph{Error Accumulation} (\emph{Drift}). In the evaluation, our approach combined with 6 popular optical flow algorithms and show significant improvement against baselines methods on 7 benchmark sequences. Such datasets include 6 synthetic benchmark sequences with realworld deformation and 1 realworld sequence.

\vspace{-3mm}
\section{Acknowledgements}
\vspace{-2mm}

We thank Ravi Garg and Lourdes Agapito for providing their GT datasets. We also thank Gabriel Brostow and the UCL Vision Group for their generous comments. The authors are supported by the EPSRC CDE EP/L016540/1 and CAMERA EP/M023281/1; and EPSRC projects EP/K023578/1 and EP/K02339X/1.

\vspace{-5mm}

\bibliographystyle{plainnat}
\bibliography{bib}

\begin{thebibliography}{28}
\providecommand{\natexlab}[1]{#1}
\providecommand{\url}[1]{\texttt{#1}}
\expandafter\ifx\csname urlstyle\endcsname\relax
  \providecommand{\doi}[1]{doi: #1}\else
  \providecommand{\doi}{doi: \begingroup \urlstyle{rm}\Url}\fi

\bibitem[Baker et~al.(2011)Baker, Scharstein, Lewis, Roth, Black, and
  Szeliski]{Middlebury}
Simon Baker, Daniel Scharstein, J.~Lewis, Stefan Roth, Michael Black, and
  Richard Szeliski.
\newblock A database and evaluation methodology for optical flow.
\newblock \emph{International Journal of Computer Vision (IJCV'11)},
  92:\penalty0 1--31, 2011.
\newblock ISSN 0920-5691.

\bibitem[Beeler et~al.(2011)Beeler, Hahn, Bradley, Bickel, Beardsley, Gotsman,
  Sumner, and Gross]{Beeler}
T.~Beeler, F.~Hahn, D.~Bradley, B.~Bickel, P.~A. Beardsley, C.~Gotsman, R.~W.
  Sumner, and M.~H. Gross.
\newblock High-quality passive facial performance capture using anchor frames.
\newblock \emph{ACM Transactions on Graphics (TOG'11)}, 30\penalty0
  (4):\penalty0 75, 2011.

\bibitem[Black and Anandan(1996)]{BA}
M.J. Black and P.~Anandan.
\newblock The robust estimation of multiple motions: Parametric and
  piecewise-smooth flow fields.
\newblock \emph{Computer vision and image understanding (CVIU'96)}, 63\penalty0
  (1):\penalty0 75--104, 1996.

\bibitem[Borshukov et~al.(2005)Borshukov, Piponi, Larsen, Lewis, and
  Tempelaar-Lietz]{Borshukov}
G.~Borshukov, D.~Piponi, O.~Larsen, JP~Lewis, and C.~Tempelaar-Lietz.
\newblock Universal capture: image-based facial animation for the matrix
  reloaded.
\newblock In \emph{ACM SIGGRAPH'05 Courses}, page~16. ACM, 2005.

\bibitem[Bradley et~al.(2010)Bradley, Heidrich, Popa, and Sheffer]{Bradley}
D.~Bradley, W.~Heidrich, T.~Popa, and A.~Sheffer.
\newblock High resolution passive facial performance capture.
\newblock \emph{ACM Transactions on Graphics (TOG'10)}, 29\penalty0
  (4):\penalty0 41, 2010.

\bibitem[Brox and Malik(2011)]{Brox_Large}
T.~Brox and J.~Malik.
\newblock Large displacement optical flow: Descriptor matching in variational
  motion estimation.
\newblock \emph{IEEE Transactions on Pattern Analysis and Machine Intelligence
  (PAMI'11)}, 33:\penalty0 500--513, 2011.
\newblock ISSN 0162-8828.

\bibitem[DeCarlo and Metaxas(1996)]{DeCarlo}
D.~DeCarlo and D.~Metaxas.
\newblock The integration of optical flow and deformable models with
  applications to human face shape and motion estimation.
\newblock In \emph{Computer Vision and Pattern Recognition (CVPR'96)}, pages
  231--238, 1996.

\bibitem[Drulea and Nedevschi(2011)]{CLG-TV1}
M.~Drulea and S.~Nedevschi.
\newblock Total variation regularization of local-global optical flow.
\newblock In \emph{Intelligent Transportation Systems (ITSC'11)}, pages
  318--323. IEEE, 2011.

\bibitem[Fulkerson and Soatto(2012)]{GPU_SIFT}
Brian Fulkerson and Stefano Soatto.
\newblock Really quick shift: Image segmentation on a gpu.
\newblock In \emph{Trends and Topics in Computer Vision}, pages 350--358.
  Springer, 2012.

\bibitem[Garg et~al.(2013{\natexlab{a}})Garg, Roussos, and Agapito]{Garg}
R.~Garg, A.~Roussos, and L.~Agapito.
\newblock A variational approach to video registration with subspace
  constraints.
\newblock \emph{International journal of computer vision (IJCV'13)},
  104\penalty0 (3):\penalty0 286--314, 2013{\natexlab{a}}.

\bibitem[Garg et~al.(2013{\natexlab{b}})Garg, Roussos, and Agapito]{Garg13}
Ravi Garg, Anastasios Roussos, and Lourdes Agapito.
\newblock Dense variational reconstruction of non-rigid surfaces from monocular
  video.
\newblock In \emph{IEEE Conference on Computer Vision and Pattern Recognition
  (CVPR'13)}, pages 1272--1279, 2013{\natexlab{b}}.

\bibitem[Godard et~al.(2015)Godard, Hedman, Li, and Brostow]{reflection}
Clement Godard, Peter Hedman, Wenbin Li, and Gabriel~J Brostow.
\newblock Multi-view reconstruction of highly specular surfaces in uncontrolled
  environments.
\newblock In \emph{International Conference on 3D Vision (3DV'15)}, pages
  19--27. IEEE, 2015.

\bibitem[Horn and Schunck(1981)]{HS}
B.K.P. Horn and B.G. Schunck.
\newblock Determining optical flow.
\newblock \emph{Artificial intelligence}, 17\penalty0 (1-3):\penalty0 185--203,
  1981.

\bibitem[Li(2013)]{li2013nonrigid}
Wenbin Li.
\newblock \emph{Nonrigid Surface Tracking, Analysis and Evaluation}.
\newblock PhD thesis, University of Bath, 2013.

\bibitem[Li et~al.(2012)Li, Cosker, and Brown]{APO}
Wenbin Li, Darren Cosker, and Matthew Brown.
\newblock An anchor patch based optimisation framework for reducing optical
  flow drift in long image sequences.
\newblock In \emph{Asian Conference on Computer Vision (ACCV'12)}, pages
  112--125, November 2012.

\bibitem[Li et~al.(2013)Li, Cosker, Brown, and Tang]{LME}
Wenbin Li, Darren Cosker, Matthew Brown, and Rui Tang.
\newblock Optical flow estimation using laplacian mesh energy.
\newblock In \emph{IEEE Conference on Computer Vision and Pattern Recognition
  (CVPR'13)}, pages 2435--2442. IEEE, June 2013.

\bibitem[Li et~al.(2014)Li, Chen, Lee, Ren, and Cosker]{moBlur}
Wenbin Li, Yang Chen, JeeHang Lee, Gang Ren, and Darren Cosker.
\newblock Robust optical flow estimation for continuous blurred scenes using
  rgb-motion imaging and directional filtering.
\newblock In \emph{IEEE Winter Conference on Application of Computer Vision
  (WACV'14)}, pages 792--799. IEEE, March 2014.

\bibitem[Li et~al.(2016)Li, Chen, Lee, Ren, and Cosker]{moblur_nc}
Wenbin Li, Yang Chen, JeeHang Lee, Gang Ren, and Darren Cosker.
\newblock Blur robust optical flow using motion channel.
\newblock \emph{Neurocomputing}, 0\penalty0 (0):\penalty0 12, 2016.

\bibitem[Lowe(2004)]{SIFT}
D.G. Lowe.
\newblock Distinctive image features from scale-invariant keypoints.
\newblock \emph{International journal of computer vision (IJCV'04)},
  60\penalty0 (2):\penalty0 91--110, 2004.

\bibitem[Lv et~al.(2013)Lv, Tek, Da~Silva, Empereur-Mot, Chavent, and
  Baaden]{lv2013game}
Zhihan Lv, Alex Tek, Franck Da~Silva, Charly Empereur-Mot, Matthieu Chavent,
  and Marc Baaden.
\newblock Game on, science-how video game technology may help biologists tackle
  visualization challenges.
\newblock \emph{PloS one}, 8\penalty0 (3), 2013.

\bibitem[Lv et~al.(2014)Lv, Halawani, Feng, Li, and
  R{\'e}hman]{lv2014multimodal}
Zhihan Lv, Alaa Halawani, Shengzhong Feng, Haibo Li, and Shafiq~Ur R{\'e}hman.
\newblock Multimodal hand and foot gesture interaction for handheld devices.
\newblock \emph{ACM Transactions on Multimedia Computing, Communications, and
  Applications (TOMM)}, 11\penalty0 (1):\penalty0 10, 2014.

\bibitem[Pizarro and Bartoli(2012)]{Pizarro}
Daniel Pizarro and Adrien Bartoli.
\newblock Feature-based deformable surface detection with self-occlusion
  reasoning.
\newblock \emph{International Journal of Computer Vision (IJCV'12)},
  97:\penalty0 54--70, 2012.
\newblock ISSN 0920-5691.

\bibitem[Salzmann et~al.(2007)Salzmann, Hartley, and Fua]{Salzmann}
M.~Salzmann, R.~Hartley, and P.~Fua.
\newblock Convex optimization for deformable surface 3-d tracking.
\newblock In \emph{International Conference on Computer Vision (ICCV'07)},
  pages 1--8, 2007.

\bibitem[Sun et~al.(2010)Sun, Roth, and Black]{Sun10}
Deqing Sun, S.~Roth, and M.J. Black.
\newblock Secrets of optical flow estimation and their principles.
\newblock In \emph{Computer Vision and Pattern Recognition (CVPR'10)}, pages
  2432--2439, 2010.

\bibitem[Tang et~al.(2012)Tang, Cosker, and Li]{tang}
Rui Tang, Darren Cosker, and Wenbin Li.
\newblock Global alignment for dynamic 3d morphable model construction.
\newblock In \emph{Workshop on Vision and Language (V\&LW'12)}, 2012.

\bibitem[Wedel et~al.(2009)Wedel, Pock, Zach, Bischof, and Cremers]{ITV_L1}
A.~Wedel, T.~Pock, C.~Zach, H.~Bischof, and D.~Cremers.
\newblock An improved algorithm for tv-l 1 optical flow.
\newblock In \emph{Statistical and Geometrical Approaches to Visual Motion
  Analysis}, pages 23--45. Springer, 2009.

\bibitem[White et~al.(2007)White, Crane, and Forsyth]{White}
R.~White, K.~Crane, and D.A. Forsyth.
\newblock Capturing and animating occluded cloth.
\newblock \emph{ACM Transactions on Graphics (TOG'07)}, 26\penalty0
  (3):\penalty0 34, 2007.

\bibitem[Yang et~al.(2015)Yang, Lin, Gao, Lv, Wei, and Song]{yang2015quality}
Jiachen Yang, Yancong Lin, Zhiqun Gao, Zhihan Lv, Wei Wei, and Houbing Song.
\newblock Quality index for stereoscopic images by separately evaluating adding
  and subtracting.
\newblock \emph{PloS one}, 10\penalty0 (12), 2015.

\end{thebibliography}

\end{document}